\pdfoutput=1

\documentclass[11pt]{article}

\usepackage[final]{coling}

\usepackage{times}
\usepackage{latexsym}

\usepackage[T1]{fontenc}

\usepackage[utf8]{inputenc}

\usepackage{microtype}

\usepackage{inconsolata}

\usepackage{graphicx}

\usepackage{amsfonts,amssymb,amsmath} 
\usepackage{booktabs}
\usepackage{multirow}
\usepackage{xspace}
\newcommand{\themodel}{GraphOTTER\xspace}
\usepackage{makecell}
\usepackage{url}
\usepackage{tcolorbox}
\tcbuselibrary{raster}
\tcbuselibrary{breakable}
\usepackage{enumitem}

\usepackage{newtxtext }  
\usepackage{listings}
\lstdefinestyle{demo}{
    basicstyle=\fontsize{8}{10}\ttfamily,
    keywordstyle=\color{blue},
    commentstyle=\color{gray},
    stringstyle=\color{green},
    showstringspaces=false,
    breaklines=true,
    breakatwhitespace=false,
    breakindent=0pt,
    escapeinside={(*@}{@*)},
    literate={á}{{\'a}}1 {ã}{{\~a}}1 {é}{{\'e}}1,
}
\usepackage{xcolor} 

\usepackage{fvextra}

\lstdefinelanguage{json}{
    basicstyle=\small\ttfamily,
    numbers=left,
    numberstyle=\scriptsize,
    stepnumber=1,
    numbersep=8pt,
    showstringspaces=false,
    breaklines=true,
    frame=lines,
    backgroundcolor=\color{white},
    literate=
      {:}{{{\color{purple}{:}}}}{1}
      {,}{{{\color{purple}{,}}}}{1}
      {\{}{{{\color{brown}{\{}}}}{1}
      {\}}{{{\color{brown}{\}}}}}{1}
      {[}{{{\color{brown}{[}}}}{1}
      {]}{{{\color{brown}{]}}}}{1},
}

\usepackage{algorithm}
\usepackage{algorithmic}

\title{GraphOTTER: Evolving LLM-based Graph Reasoning for \\Complex Table Question Answering}

\author{
Qianlong Li$^{\spadesuit\heartsuit}$, \quad
Chen Huang$^{\spadesuit\heartsuit}$, \quad
Shuai Li$^{\spadesuit\heartsuit}$ \\
\textbf{Yuanxin Xiang}$^{\spadesuit\heartsuit}$, \quad 
\textbf{Deng Xiong}$^{\clubsuit}$, \quad 
\textbf{Wenqiang Lei}$^{\spadesuit\heartsuit}$\thanks{Corresponding author.}
\\
${\spadesuit}$ Sichuan University \quad ${\clubsuit}$ Stevens Institute of Technology \\ 
${\heartsuit}$ Engineering Research Center of Machine Learning and Industry Intelligence,\\Ministry of Education, China \\
\{liqianlong0922, huangc.scu, lsbeatsleo, whutxyx\}@gmail.com \\
dxiong@stevens.edu \quad wenqianglei@scu.edu.cn 
}

\begin{document}
\maketitle
\begin{abstract}
Complex Table Question Answering involves providing accurate answers to specific questions based on intricate tables that exhibit complex layouts and flexible header locations. Despite considerable progress having been made in the LLM era, the reasoning processes of existing methods are often implicit, feeding the entire table into prompts, making it difficult to effectively filter out irrelevant information in the table. To this end, we propose \themodel that explicitly establishes the reasoning process to pinpoint the correct answers. In particular, \themodel leverages a graph-based representation, transforming the complex table into an undirected graph. It then conducts step-by-step reasoning on the graph, with each step guided by a set of pre-defined intermediate reasoning actions. As such, it constructs a clear reasoning path and effectively identifies the answer to a given question. Comprehensive experiments on two benchmark datasets and two LLM backbones demonstrate the effectiveness of \themodel. Further analysis indicates that its success may be attributed to the ability to efficiently filter out irrelevant information, thereby focusing the reasoning process on the most pertinent data. Our code and experimental datasets are available at \url{https://github.com/JDing0521/GraphOTTER}.
\end{abstract}

\section{Introduction}
\begin{figure}
    \centering
    \includegraphics[width=0.5\textwidth]{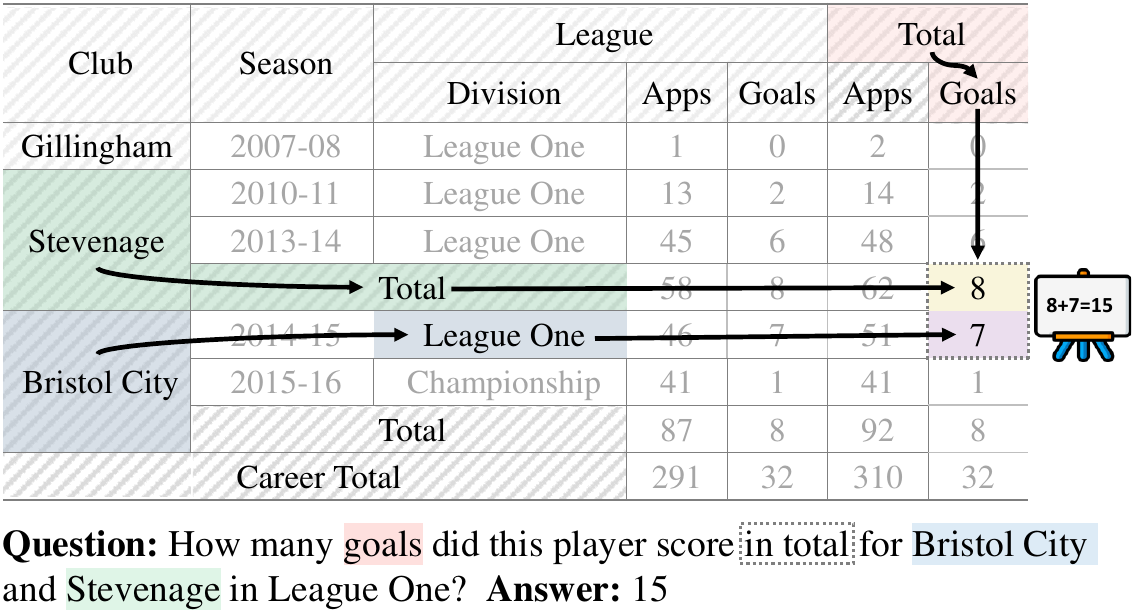}
    \caption{Example of complex table QA. Header cells are styled with a dashed background, while the reasoning process is highlighted using color.}
    \label{fig:complex}
    \vspace{-5mm}
\end{figure}

Complex tables in documents frequently employ advanced layouts like merged cells and flexible header locations \cite{zheng2023tqa}. This complexity facilitates the structured presentation of detailed information in a flexible manner, making them a common feature in financial reports \cite{zhu2021tatqa, chen2021finqa} and other professional documents \cite{wang2021scientificqa, zhong2020imagetablereco}. In this context, \textbf{Complex Table Question Answering} (QA) recently emerges as a crucial task \cite{zhao2023tableparser, zheng2023tqa, E5}, enabling the extraction of valuable information from complex tables. However, merged cells and nested structures contained in the complex tables introduce implicit semantic relationships between entities within cells, posing significant challenges for table structure understanding \cite{katsis2021aitqa, cheng2021hitab}. Taking Figure \ref{fig:complex} for example, a "total" cell represents scores across different divisions for the Bristol City, while for the Stevenage, it only encompasses scores from a single division.

To address the challenges, current research focuses on prompting large language models (LLMs) to to implicitly reason and derive answers based on carefully designed table representations. In particular, they propose transforming the table into more accessible formats such as tuples \cite{zhao2023tableparser}, Markdown \cite{fewshot,liu2023mixsc}, and HTML \cite{E5}. Some of these approaches also involve annotated table headers within the transformed representations to further improve the model's understanding of complex, hierarchical structures \cite{zhao2023tableparser}. Subsequently, these techniques integrate the transformed table representations into carefully designed prompts to unlock the reasoning potential of LLMs \cite{cheng2021hitab, zhao2023tableparser, E5}, thereby enabling them to pinpoint accurate answers within complex tables.

However, the reasoning processes employed by existing approaches are frequently embedded within simple, free-form prompts, which struggles to offer a structured way to represent the intricate relationships among data in different (merged) columns and rows. As illustrated in Figure \ref{fig:complex}, identifying the correct answer typically involves focusing on a limited number of cells within the table (highlighted in color). In such scenarios, an implicit reasoning process that relies on the entire table representation might not effectively filter out irrelevant information (marked in white).
This limitation can significantly diminish the effectiveness of complex table question answering (QA) systems and reduce their practical utility when dealing with detailed and intricate tabular data.

To this end, we aim to explicitly establishing the reasoning process for complex table QA, thereby assisting LLMs in filtering out irrelevant information and enhancing the task effectiveness. In this paper, we propose a novel approach, called \textbf{\themodel}, that evolves \underline{Graph} reasoning for c\underline{O}mplex \underline{T}able ques\underline{T}ion answ\underline{ER}ing. 
In particular, \themodel leverages a graph-based representation, transforming the complex table into an undirected graph where header and data cells are treated as nodes. It also introduces a set of intermediate reasoning actions on the graph, such as \textit{VisitNode}, \textit{GetSharedNeighbours}, and \textit{AnswerQuestion}, which act as tools to facilitate navigation and reasoning within the graph structure. Building upon this, \themodel prompts LLMs to conduct step-by-step reasoning on the graph. For each reasoning step, it selects the most appropriate action to guide the LLM towards the answer. To facilitate the selection, a reasoning trace is maintained to track visited nodes, enabling \themodel to effectively steer the reasoning process and filter out irrelevant information. As such, it constructs a clear reasoning path and effectively identifies the answer.

To evaluate our effectiveness, we conduct comparative experiments with various baselines using two benchmark datasets and two LLM backbones\footnote{Taking their costs and effectiveness into consideration.}. Compared to baselines, \themodel achieves a notable improvement in complex table question answering, exhibiting an average performance gain of +4.77\% over the best baseline. Our in-depth analysis reveals that this success can be attributed to the explicit reasoning process, which reduce irrelevant information in the graph/table. Moreover, we experimentally show that combining graph representations with explicit reasoning creates a powerful synergy that harnesses the flexibility of the graph while mitigating its inherent complexity. This approach holds significant promise for accurately pinpointing answers from complex tables. 
To sum up, we claim the following contributions.
\begin{itemize}[leftmargin=*,itemindent=0.05cm, itemsep=-4pt]
    \item We call attention to the importance of explicitly establishing the reasoning process for Complex Table QA. This promotes the task effectiveness and its real-world utility.
    \item We propose \themodel, a novel LLM-based approach that evolves graph reasoning to construct a clear reasoning path and effectively identifies the answer to a given question.
    \item We conduct extensive experiments on benchmark datasets to show the superiority of \themodel. We further analyze the potential reason for the effectiveness of graph reasoning.
\end{itemize}

\section{Related work}
We focus on evolving LLM-based graph reasoning for complex table QA task. We clarify our difference to existing methods as follows.

\noindent\textbf{Representation for Complex Table}. 
Unlike simple tables \cite{zhu2021tat, deng2022pacific}, which can be readily processed, the complex structures of real-world tables necessitate the development of more sophisticated representations to facilitate effective model comprehension.
This includes transforming table headers into hierarchy-aware logical form \cite{cheng2021hitab} and converting the entire table into tuples \cite{zhao2023tableparser}, Markdown \cite{fewshot,liu2023mixsc}, and HTML \cite{E5}. However, these methods typically rely on annotated table headers or assume that headers are located only at the top or left side of the table. 
In contrast, \themodel empowers the LLM to infer header cells among connected nodes in the graph, enhancing its adaptability to diverse table structures.

\begin{figure*}[t]
    \centering
    \includegraphics[width=0.97\textwidth]{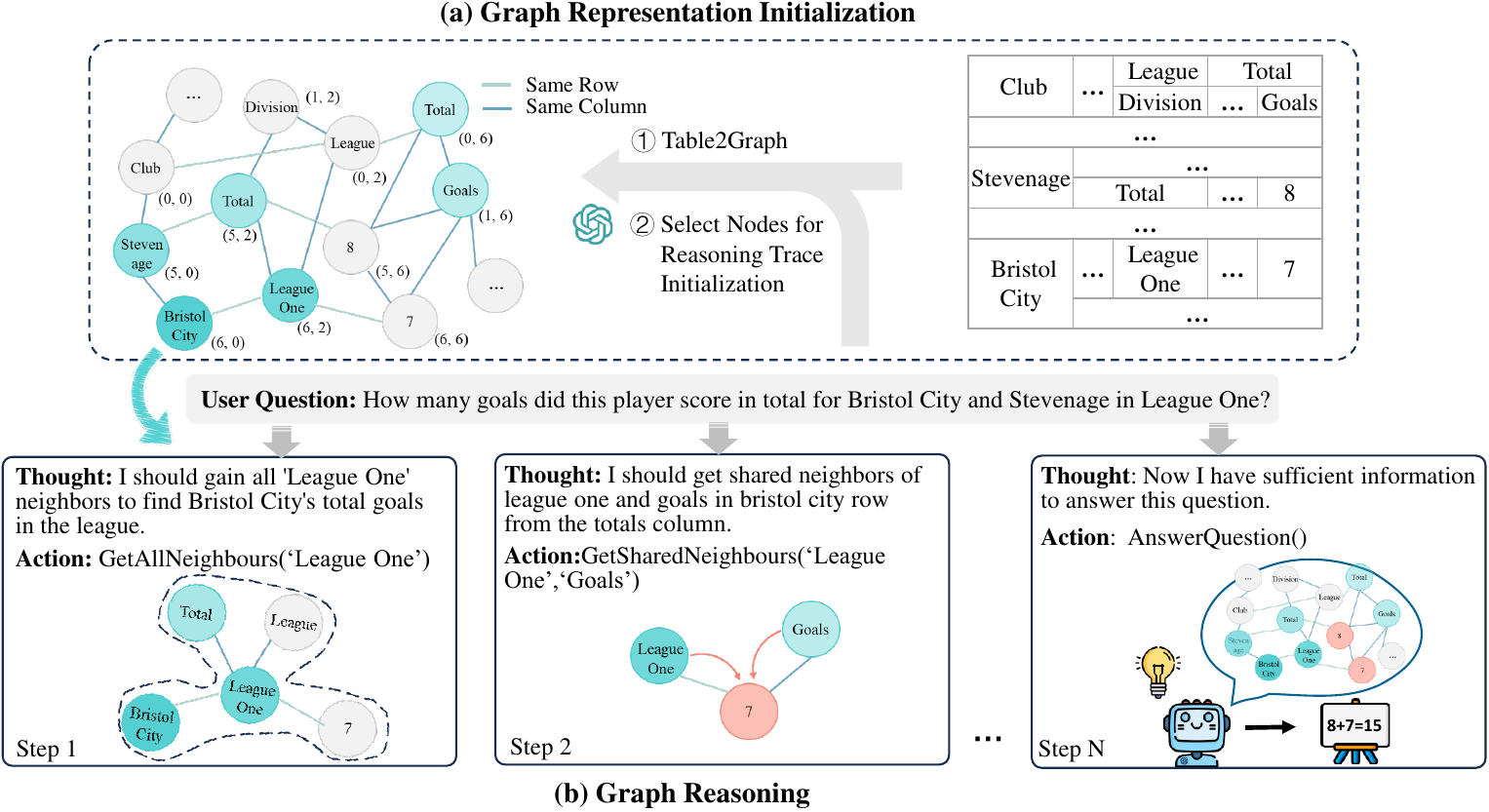}
    
    \caption{
    Overview of \themodel. It first transforms the table into a graph representation (For clarity, this graph only shows some of the edges between nodes in the same row/column). Then, it establishes a step-by-step reasoning process on this graph, guided by a reasoning trace and a set of pre-defined intermediate actions (Note that the nodes in the reasoning trace are colored). These actions facilitate navigation and reasoning within the graph structure, leading to a more explicit and efficient approach for complex table QA.
    }
    \label{fig:overview}
\end{figure*}

\noindent\textbf{Reasoning for Complex Table QA}.
Complex table QA relies on the ability to reason with information directly from the table itself to derive accurate answers to specific questions. Currently, the reasoning processes of existing methods employed are often implicitly represented within free-form text prompts \cite{cheng2021hitab, zhao2023tableparser}, fine-tuned table embeddings based on the GCN or transformer \cite{zheng2023tqa, jia2023getpt}, or code interpreter \cite{E5}, which struggles to offer a structured way to represent the intricate relationships among data in different (merged) columns and rows. In this case, we argue that the implicit reasoning may be inadequate when being tasked with reasoning over complex tables. By explicitly representing the reasoning process, \themodel overcomes the limitations, paving the way for more accurate complex table QA. Notably, \themodel distinguishes itself from existing graph-based complex table QA methods, which typically rely on fine-tuning transformers and GNN-style models for table structure encoding \cite{zheng2023tqa, jia2023getpt}. \themodel, in contrast, explicitly reasons through a series of intermediate steps on the graph via in-context learning. Also, \themodel may be conceptually close to Chain-of-Table \cite{wang2024chainofable} yet vitally different due to the following reasons. While Chain-of-Table focuses on simple table understanding, employing multi-step tabular reasoning to form a chain of transformed tables, \themodel leverages graph reasoning to address the complex table QA. See Table \ref{tab:main results} for our effectiveness.

\textbf{LLM-based Graph Reasoning}. 
LLM-based graph reasoning  involves utilizing LLMs to analyze and interpret the graph, extracting new insights or relationships based on the graph's structure and properties \cite{agrawal2024llmsperformstructuredgraph}. This process typically entails a multi-step approach: identifying the required information, searching for relevant data within the graph, and generating an answer based on the retrieved information \cite{ren2024llmforgraphsurvey,shang2024llmforgraphsurvey2}. To enhance reasoning capabilities, some approaches further integrate external tools like retrievers \cite{sun2023thinkongraph,wang2023boosting,jiang2024kgagent,wu2023retrirewriteans}. 
In this paper, we evolving the LLM-based graph reasoning to solve the complex table QA task, which guides the reasoning process in a step-by-step manner, avoiding irrelevant information.

\section{\themodel for
Complex Table QA}

\textbf{Task Formulation}. 
Given a complex table $T$ and a user question $Q$, we transform the table into a graph $G=\{\mathcal{N}, \mathcal{E}, \mathcal{V}\}$, where $\mathcal{N}$, $\mathcal{E}$ and $\mathcal{V}$ represent the sets of nodes, edges, and corresponding cell values, respectively. Leveraging this graph representation, we conduct step-by-step reasoning to pinpoint the correct answer $A$. In this process, we dynamically maintain a reasoning trace $Tr$, represented as a subgraph that consists of the visited nodes.

\subsection{Method Overview} 
\themodel aims to establish the step-by-step reasoning process for complex table QA. As depicted in Figure \ref{fig:overview}, at each reasoning step, \themodel prompts the LLM to select a reasoning action from a pre-defined set of intermediate reasoning actions $\mathcal{R}$ (detailed in Table \ref{tab:functions}). 
To guide this process, \themodel introduces a reasoning trace $Tr$ that tracks the nodes visited during reasoning, which further contributes to the selection of the next intermediate action. Finally, the process terminates when \themodel selects the \textit{AnswerQuestion} action, producing the final answer. See the pseudo code in Algorithm \ref{algo:overview} and the case study in Appendix \ref{case} for better understanding.

\subsection{Graph Representation Initialization} \label{table2graph} 
\themodel transforms the input table $T$ into an undirected graph $G$, where edges connect cells within the same row or column. In particular, each node $\mathcal{N}_i \in \mathcal{N}$ is denoted as a triple $(Rid, Cid, \mathcal{V}_i)$, where $Rid$ and $Cid$ are the row and column indices in the table. Notably, for merged cells, their $Rid$ or/and $Cid$ values are represented as sets containing the indices of the corresponding split cells. 
Moreover, each edge $\mathcal{E}_{ij} \in \mathcal{E}$ is denoted as a triple $(\mathcal{N}_i, \mathcal{N}_j, M_{ij})$, where $M_{ij}$ indicates whether the two connected nodes are in the same row or the same column in the table. 
Notably, both header and data cells are treated as nodes without distinction. This approach eliminates the reliance on pre-defined headers, unlike existing methods that require pre-annotated headers or assume headers are located only at the top or left side of the table \cite{wang2024chainofable,liu2023mixsc,ye2023dater,cheng2021hitab}. This enhances \themodel's adaptability to diverse table structures. 

To initiate the reasoning process, \themodel instructs the LLM to select a small set of nodes relevant to the user question $Q$ as the initial set for initializing the reasoning trace $Tr$. For implementations, we initially filter up to eight most relevant cells, with the sensitivity analysis being explored in Section \ref{hyper}.

\begin{table}[t]
\centering
    \resizebox{0.45\textwidth}{!}{
        
        \normalsize
        \begin{tabular}{p{3.2cm}p{6cm}}
            \hline
            \toprule
            \textbf{Reasoning Actions} & \textbf{Description}                                                                     \\ \midrule
            \textit{VisitNode}          & \begin{tabular}[c]{@{}p{6cm}@{}} Go to the specified node on the graph.\end{tabular}       \\
            \textit{GetAllNeighbours}    & \begin{tabular}[c]{@{}p{6cm}@{}}Get all neighbors of the specified node.\end{tabular} \\
            \textit{GetSharedNeighbours} & \begin{tabular}[c]{@{}p{6cm}@{}} Get all shared neighbors between two specified nodes.\end{tabular}          \\
            \textit{AnswerQuestion}  & \begin{tabular}[c]{@{}p{6cm}@{}}Answer the user question.\end{tabular} \\ \bottomrule
        \end{tabular}
}
\caption{Reasoning actions on the graph. They are simple yet flexible, encompassing common operations required for reasoning on the graph.}
\label{tab:functions}
\vspace{-3mm}
\end{table}

\subsection{Graph Reasoning for Complex Table QA}
\label{graphreasoning} 
\themodel implements a \textbf{Thought-Action-Update} process for step-by-step graph reasoning. To guide the reasoning process, it further introduces a reasoning trace, $Tr$, which tracks visited nodes during reasoning. Specifically, at each step, \themodel prompts the LLM to first think about the current reasoning state based on the current reasoning trace and other relevant arguments. Based on this thought process, \themodel selects an action from the reasoning action set $\mathcal{R}$, as outlined in Table \ref{tab:functions}. Finally, the trace is updated after \themodel executes the chosen action, which in turn influences the subsequent reasoning state.

\textbf{Thought}. \themodel gathers scattered information to establish its current reasoning state, considering not only the global information provided by the entire graph but also local information specific to the input question. This local information includes the outputs of the previous \textit{Thought} and \textit{Action} step, the historical reasoning trace, which helps distill key insights from past reasoning steps, allowing \themodel to focus on important details, and self-inferred semantic relationships between connected nodes, leveraging the LLM's semantic understanding capabilities to identify header cells.

\textbf{Action}. We guide \themodel to select the subsequent intermediate action from the action set $R$ based on the reasoning state. As outlined in Table \ref{tab:functions}, we have designed a simple yet flexible set of reasoning actions encompassing common operations required for reasoning on the graph. These actions, acting as tools, enhance \themodel's reasoning performance. In particular, Action \textit{VisitNode} facilitates node selection and traversal within the graph. In scenarios where multiple nodes share the same value, \themodel presents these nodes to the LLM for selection, accompanied by structural information ($Cid$ and $Rid$) to enable informed decision-making. Furthermore, actions \textit{GetAllNeighbours} and \textit{GetSharedNeighbours} empower \themodel to focus on local information and analyze the surrounding neighbors of selected node(s). We also incorporate the action of answering questions within this set. This aligns with the concept of learning to defer \cite{madras2018predict, tailor2024learning}, which promotes \themodel to automatically evaluate the need for additional information before answering, rather than relying on heuristic rules as in \cite{wang2024chainofable}. 

Notably, unlike implicit reasoning where the entire table is embedded in a prompt \cite{cheng2021hitab, zhao2023tableparser,E5}, our method explicitly establishes the reasoning path via executing these actions, which could effectively filter out irrelevant information and guide reasoning towards a more accurate answer, a topic we will analysis in Section \ref{depth analysis}.

\textbf{Update}. The reasoning trace $Tr$ is updated by appending the new visited node after \themodel executes the \textit{VisitNode} action. The newly added node establish edges with previously visited nodes if they share common neighbors or are directly connected in the original graph. This process effectively updates the local information for the subsequent \textit{Thought} step. In this way, \themodel forms a iterative reasoning on the graph to pinpoint the correct answer.

\begin{algorithm}
\caption{Pseudo code of \themodel}
\label{algo:overview}

    \begin{algorithmic}[1] 
    \ENSURE $(T, Q)$ is a complex table-question pair.

    \STATE $\triangleright$ \textit{See Section \ref{table2graph} for details.}  
    \STATE Transform the table $T$ into the graph $G$.
    \STATE Initialize reasoning trace $Tr$ based on $G$.
    \STATE $\triangleright$ \textit{Reasoning on the graph, see Section \ref{graphreasoning}.} 

    \REPEAT
        \STATE Update reasoning state via \textbf{Thought} based on reasoning trace $Tr$
        \STATE Reason one step via \textbf{Action} based the state
        \STATE Modify reasoning trace $Tr$ via \textbf{Update}
    \UNTIL{\textit{AnswerQuestion} is selected}
    
    \STATE $\triangleright$ \textit{See Section \ref{answer} for details.} 
    \STATE Answer generation via in-context learning.
    \end{algorithmic}
\end{algorithm}

\subsection{Answer Generation}
\label{answer}
When the \textit{AnswerQuestion} action is selected, we instruct the LLM via CoT prompts to produce the final answer to the user question based on the question and current local reasoning state from \textit{Thought} step, instead of using the whole graph.

\section{Experiments}

To evaluate our effectiveness, we conduct comparative experiments with various baselines using two benchmark datasets and two LLM backbones. In particular, we study how effective is our \themodel compare to existing complex table QA methods (Section \ref{mainm}). Furthermore, we comprehensively analyze the advantages and characteristics of \themodel in Section \ref{depth analysis} and \ref{hyper}. For better understanding the reasoning process of \themodel, we report case studies in Appendix \ref{case} due to the space limit. Finally, our code and experimental datasets are available at \url{https://github.com/JDing0521/GraphOTTER}.

\subsection{Experimental Setups}

\textbf{Dataset}. Following \citet{zhao2023tableparser}, we use two benchmark datasets: Hitab\cite{cheng2021hitab} and AIT-QA\cite{katsis2021aitqa}. These two datasets contain a large number of complex tables collected from industrial documents, which not only provide larger-scale tables but also reflect the real data analysis scenarios in the industry. Statistics of the datasets can be found in Appendix \ref{app:dataset statistics}.

\noindent\textbf{Baselines}.
Current approaches to complex table QA rely on predefined table headers and implicit reasoning processes. These methods can be broadly categorized into two types: \textit{One-shot Implicit Reasoning} and \textit{Iterative Implicit Reasoning}, which prompts LLMs in a iterative manner. We also include the only existing method that employs \textit{iterative explicit reasoning} as a baseline.
\begin{itemize}[leftmargin=*]
    \item \textbf{One-shot Implicit Reasoning}. This category encompasses TableReasoner \cite{fewshot} and TableParser \cite{zhao2023tableparser}, which leverage carefully crafted (CoT) prompts and table representations based on Markdown or tuples.
    
    \item \textbf{Iterative Implicit Reasoning}. ReAct \cite{react} utilizes HTML-based table representations and guides LLMs in generating (Thought, Act, Observation) tuples. E$^{5}$\cite{E5} builds upon ReAct, further incorporating self-described table structure information and a code interpreter to enhance the action generation process. MIX-SC \cite{liu2023mixsc} employs multiple reasoning attempts for self-consistency, including carefully designed DataFrame-based representation and ReAct-style iterative reasoning using Python interpreter and in-context learning.
    \item \textbf{Iterative Explicit Reasoning}. This involves Chain-of-Table \cite{wang2024chainofable}. While it is designed for simple tables, it stands out as the only iterative explicit reasoning method. It utilizes multi-step reasoning on the table, forming a chain of transformed tables through tabular operations, rather than our graph reasoning.
\end{itemize}

\noindent\textbf{Evaluation Metrics}. Following previous research \cite{zheng2023tqa,fewshot,E5,liu2023mixsc, zhao2023tableparser}, we employ a multifaceted evaluation approach. Firstly, we utilize \underline{Exact Match (EM)} to quantify the precision of model outputs against the ground truth. Furthermore, we leverage a \underline{LLM-based evaluator (LLM Eval)} to assess the accuracy of predictions. The detailed implementation is offered in Appendix \ref{tas}.

\begin{table*}[]
    \centering
    \renewcommand\arraystretch{1.2}
    \resizebox{1.0\textwidth}{!}{%
        \begin{tabular}{clcccccccc}
    \hline
    \toprule
    
    \multirow{3}{*}{\textbf{Reasoning Type}} &
      \multicolumn{1}{c}{\multirow{3}{*}{\textbf{Methods}}} &
      \multicolumn{4}{c}{\textbf{Gemini 1.5}} &
      \multicolumn{4}{c}{\textbf{Qwen2}} \\ \cline{3-10} 
     &
      \multicolumn{1}{c}{} &
      \multicolumn{2}{c}{\textbf{Hitab}} &
      \multicolumn{2}{c}{\textbf{AIT-QA}} &
      \multicolumn{2}{c}{\textbf{Hitab}} &
      \multicolumn{2}{c}{\textbf{AIT-QA}} \\ \cline{3-10} 
     &
      \multicolumn{1}{c}{} &
      \textbf{EM} &
      \textbf{LLM Eval} &
      \textbf{EM} &
      \textbf{LLM Eval} &
      \textbf{EM} &
      \textbf{LLM Eval} &
      \textbf{EM} &
      \textbf{LLM Eval} \\  \midrule
    \multirow{2}{*}{\textit{\begin{tabular}[c]{@{}c@{}}One-shot Implicit\end{tabular}}} &
      TableReasoner\cite{fewshot} &
      44.19 &
      68.18 &
      65.40 &
      77.38 &
      56.41 &
      75.68 &
      76.29 &
      88.83 \\
     &
      TableParser\cite{zhao2023tableparser} &
      57.51 &
      68.94 &
      63.22 &
       \underline{ 78.20} &
      44.57 &
      69.76 &
      64.85 &
      83.11 \\ \hline
    \multirow{3}{*}{\textit{\begin{tabular}[c]{@{}c@{}}Iterative Implicit\end{tabular}}} &
      ReAct\cite{react} &
      43.88 &
      47.60 &
      47.14 &
      50.41 &
      54.73 &
      57.89 &
      68.12 &
      72.75 \\
     &
      E5\cite{E5} &
      57.26 &
      62.44 &
      59.40 &
      64.31 &
      43.56 &
      47.16 &
      56.40 &
      59.13 \\
     &
      MIX-SC\cite{liu2023mixsc} &
      \underline{ 62.41} &
       \underline{ 68.99} &
       \underline{ 69.95} &
      74.86 &
       \underline{ 73.42} &
       \underline{ 77.08} &
       \underline{ 84.20} &
       \underline{ 91.83} \\ \hline
    \multirow{2}{*}{\textit{\begin{tabular}[c]{@{}c@{}}Iterative Explicit\end{tabular}}} &
      Chain-of-Table\cite{wang2024chainofable} &
      25.82 &
      36.68 &
      18.53 &
      23.16 &
      44.26 &
      62.69 &
      49.32 &
      61.04 \\ 
     &
      \themodel (\textit{Ours}) &
      \textbf{67.76} &
      \textbf{69.66} &
      \textbf{81.47} &
      \textbf{82.56} &
      \textbf{73.74} &
      \textbf{77.37} &
      \textbf{88.28} &
      \textbf{92.64} \\ \bottomrule
    \end{tabular}%
}
\caption{Overall evaluation on complex table QA. Our \themodel stands out as the most effective method for this task across two benchmark datasets and LLM backbones.}
\label{tab:main results}
\end{table*}

\noindent\textbf{Implementation Details}.
Considering the financial constraints, we experiment on two LLM backbones for all methods: Gemini-1.5\footnote{Gemini-1.5-flash-latest} and Qwen2\footnote{Qwen2-72B-Instruct}. 
To ensure reproducibility, we maintain consistent temperature and random seed settings for both backbones (where applicable). For implementations of baselines, we rigorously adhered to their official GitHub code and original paper prompts for their implementation. For \themodel, to initiate the reasoning process, we selected up to eight of the most relevant cells for all datasets. Further implementation details regarding baselines and \themodel are presented in Appendix \ref{thisis details}.

\begin{table}[]
\renewcommand\arraystretch{1.2}
\centering
\resizebox{0.45\textwidth}{!}{%
    \begin{tabular}{lcccc}
    \toprule
    \multicolumn{1}{c}{\multirow{2}{*}{\textbf{Model}}} & \multicolumn{2}{c}{\textbf{Gemini 1.5}} & \multicolumn{2}{c}{\textbf{Qwen2}} \\ \cline{2-5} 
    \multicolumn{1}{c}{} &
      \textbf{\begin{tabular}[c]{@{}c@{}}HRQ\end{tabular}} &
      \textbf{\begin{tabular}[c]{@{}c@{}}HUQ\end{tabular}} &
      \textbf{\begin{tabular}[c]{@{}c@{}}HRQ\end{tabular}} &
      \textbf{\begin{tabular}[c]{@{}c@{}}HUQ\end{tabular}} \\ \midrule
    \multicolumn{5}{l}{\textit{One-shot Implicit Reasoning}}                                                                       \\
    TableReasoner                                       & 71.05            & 63.92            & 81.58            & 90.72           \\
    TableParser                                         & \textbf{77.63}   & 78.35            & 80.26            & 83.85           \\ \hline
    \multicolumn{5}{l}{\textit{Iterative Implicit Reasoning}}                                                                      \\
    ReAct                                               & 44.74            & 51.89            & 60.53   & 75.95           \\
    E5                                                  & 63.16            & 64.60            & 53.95            & 60.48           \\
    MIX-SC                                              & 59.21            & \underline{ 78.97}      & \textbf{84.21}   & \underline{ 93.81}     \\
    \hline
    \multicolumn{5}{l}{\textit{Iterative Explicit Reasoning}}                                   \\
    Chain-of-Table                                      & 28.95            & 21.65            & 57.89            & 61.86           \\ 
    GraphOTTER(\textit{Ours})                                    & \underline{ 69.74}      & \textbf{85.91}   & \textbf{84.21}   & \textbf{94.85}  \\ \bottomrule
    \end{tabular}%
}
\caption{Detailed evaluation on different user questions, categorized by whether their answers are related to the table headers or not. "HRQ" represents "Header-related questions", and "HUQ" represents "Header-unrelated questions". Here, LLM Eval is used. For results on EM, refer to Table \ref{tab:aitqa-EM-result} in appendix.}
\label{tab:aitqa-result}
\vspace{-2mm}
\end{table}

\subsection{Main Results}
\label{mainm}
 Table \ref{tab:main results} and Table \ref{tab:aitqa-result} present an overview of the results across two benchmark datasets and LLM backbones. These results suggest that \themodel stands out as the leading method for complex table QA. We provide further analysis below.

 \textbf{\textit{How effective is \themodel?} -- It consistently outperforms all baselines across various datasets and LLM backbones, demonstrating its superiority in complex table QA}. As illustrated in Table \ref{tab:main results}, while the MIX-SC method demonstrated improved performance by concurrently incorporating multiple inference techniques, including direct prompts and ReAct, and employing self-consistency for aggregation, its results still lag behind our proposed approach. Notably, our method surpasses MIX-SC, the best baseline, by an average of +9.07\% and +1.63\% on datasets Hitab and AIT-QA, respectively, highlighting the significant advantage of our model. Notably, such advantage contributes to our explicit graph reasoning, a topic we will delve into later in Section \ref{depth analysis}.

\textbf{\textit{How well does \themodel handle different user questions?} -- It achieves comparable and even superior performance}. Our model, \themodel, employs a graph representation where data and header cells are treated equally. This necessitates the model to infer the headers of connected nodes within the graph during the Thought step. To assess the effectiveness of this self-inferred header mechanism, we analyze performance on both header-related and header-unrelated user questions. Note that these questions are annotated within the AIT-QA dataset.
As illustrated in Table \ref{tab:aitqa-result}, \themodel achieves promising results on both question types, highlighting the efficacy of our self-inferred headers. Further ablation in Section \ref{depth analysis} further supports this observation. In our opinion, this performance can be attributed to the powerful semantic understanding capabilities of LLMs, enhancing the \themodel's adaptability to diverse table structures.

\subsection{Ablation \& In-depth Analysis} 
\label{depth analysis}
We delve into ablation studies and detailed analysis to investigate our strengths, where two ablations are involved. We provide the results of Gemini in Table \ref{tab:ablation study} and draw the following observations.
\begin{itemize}[leftmargin=*]
    \item \textit{\themodel w/ known headers} is provided with ground truth table header information, eliminating the need for the LLM to infer headers.
    \item \textit{\themodel w/ implicit reasoning} utilizes the one-shot implicit reasoning approach, directly prompting the LLM to generate the answer based on graph-based representation, rather than performing explicit graph reasoning. We refer to the TableReasoner\footnote{We opted to avoid using iterative implicit reasoning methods as they designed table-specified actions for ReAct-style iterations.} prompts by creating a graph-representation variant.
\end{itemize}

\begin{figure}[h]
    \centering
    \includegraphics[width=0.4\textwidth]{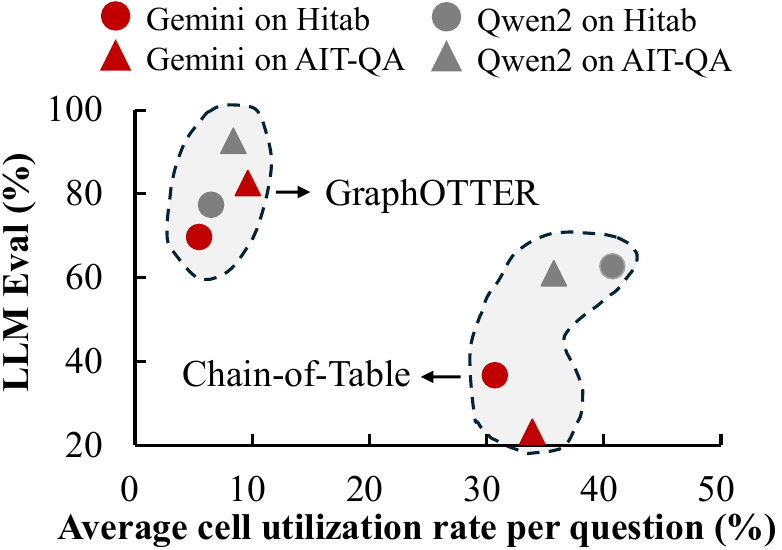}
    \caption{Average percentage of cells/nodes required for generating the final answer. We report results on explicit reasoning methods, while for implicit reasoning based methods, they use all the cells in the whole table.}
    \label{fig:cells statisics}
    \vspace{-5mm}
\end{figure}

\textbf{\textit{Why the graph reasoning is effective?} -- The success of explicit graph reasoning may be attributed to its ability to effectively filter out irrelevant information}. Compared to \themodel with its variant utilizing the same graph representation but employing the implicit reasoning, \themodel demonstrates significant improvements in both datasets. This advantage is attributed to instructing the LLM to answer the question using a specific reasoning trace rather than the entire graph/table. This effectively filters out irrelevant information from the raw table. Further analysis, as shown in Figure \ref{fig:cells statisics}, reveals that \themodel utilizes a smaller number of nodes compared to other methods when generating answers. These nodes represent the reasoning trace produced by the model's step-by-step reasoning on the graph. Based on our calculation, on average, 93.38\% of these nodes directly contain the correct answer, demonstrating the efficacy of explicit graph reasoning in eliminating irrelevant information and improving the accuracy of LLM responses.

\begin{figure*}[t]
    \centering
    \includegraphics[width=0.97\textwidth]{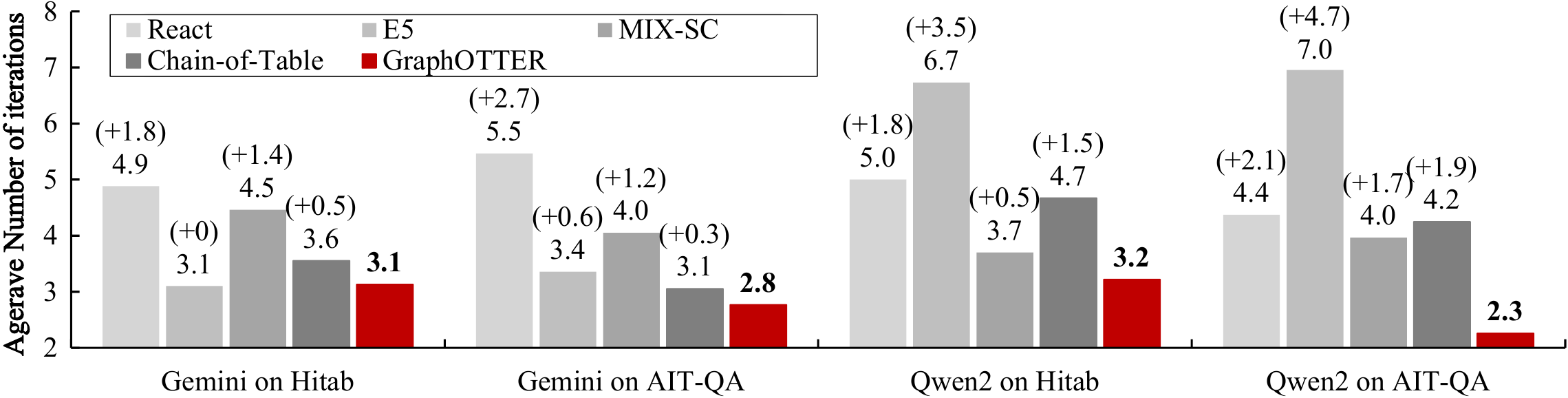}
    \caption{Illustration of the iteration efficiency of different methods. \themodel achieves superior performance while requiring fewer iterations.
    }
    \label{fig:iterations}
    \vspace{-3mm}
\end{figure*}

\textbf{\textit{How effective is the graph-based table representation?} -- Combining graphs with explicit reasoning creates a powerful synergy}. While graph representations offer significant potential, their inherent complexity (e.g., richer node relationships than tables) compared to tables can pose challenges for implicit reasoning. 

LLMs processing graphs may have difficulty distinguishing key connections from irrelevant ones, resulting in inaccuracies (Refer to \textit{TableReasoner} and the \textit{implicit reasoning variant of \themodel} in Table \ref{tab:ablation study}). 
However, combining graphs with explicit reasoning methods can alleviate these issues (cf. our results).

Explicit reasoning on the graph enables systematic exploration and identification of key connections and inferences from its structure.
This targeted approach may reduce the risk of noise or irrelevant relationships impacting results, ultimately enhancing the accuracy and reliability of the inferences. Thus, combining graphs with explicit reasoning creates a powerful synergy that harnesses the flexibility of graphs while mitigating their inherent complexity, which holds immense promise for extracting correct answer from complex table.

\textbf{\textit{How effective is the self-inferred table headers?} -- The self-inferred table headers demonstrates high availability and their effectiveness closely resemble that of using ground truth}. 
Through controlled variable experiments, we confirm the effectiveness of the self-inferred table header information, demonstrating results of \themodel comparable to those obtained using \themodel \textit{w/ known headers}.

\subsection{Characteristics of \themodel}
\label{hyper}

\begin{table}[]
    \renewcommand\arraystretch{1.2}
    \centering
    \resizebox{0.47\textwidth}{!}{
        \begin{tabular}{lcccc}
            \hline
            \toprule
            \multicolumn{1}{c}{\multirow{2}{*}{\textbf{Model}}} & \multicolumn{2}{c}{\textbf{HiTab}} & \multicolumn{2}{c}{\textbf{AIT-QA}} \\ \cline{2-5} 
            \multicolumn{1}{c}{}      & \textbf{EM}    & \textbf{LLM Eval} & \textbf{EM}    & \textbf{LLM Eval} \\ \midrule
            TableReasoner    & 44.19          & 68.18              & 76.29          & 77.38              \\
            Chain-of-Table   &25.82 &
      36.68 &
      18.53 &
      23.16  \\
            \hline
            \themodel                & \underline{67.76}          & \underline{69.66}              & \underline{81.47}          & \underline{82.56}              \\
            \textit{w/ known headers} & \textbf{68.46} & \textbf{70.55}     & \textbf{82.29} & \textbf{82.83}     \\
            
           \textit{w/ implicit reasoning} & 27.97          & 67.05              & 45.78          & 79.84              \\ 
            \bottomrule
        \end{tabular}
    }
\caption{Ablation studies using Gemini. Combining the graph with explicit reasoning creates a powerful synergy for complex table QA. The \themodel-inferred table headers demonstrates high availability.}
\label{tab:ablation study}
\vspace{-3mm}
\end{table}

This section aims to uncover characteristics of \themodel in terms of iteration efficiency and hyper-parameter sensitivity within reasoning trace initialization. Detailed observations are as follows.

\textbf{\textit{How is the iteration efficiency of \themodel?} -- \themodel achieves superior performance while requiring fewer iterations}.  
Figure \ref{fig:iterations} presents the average number of iterations per user question for \themodel and iterative reasoning-based baselines across two datasets. The results demonstrate that \themodel consistently requires fewer iterations while achieving higher QA performance (cf. Table \ref{tab:main results}). Importantly, ReAct and E5 exhibit a higher number of iterations across different backbones, primarily due to errors in the LLM-generated Python code, leading to a recursive cycle of erroneous corrections. Chain-of-Table, designed for simpler table understanding, struggles to model complex relationships between multiple cells, necessitating additional iterations to retrieve comprehensive information. In contrast, \themodel's graph-based representation and step-by-step reasoning capabilities enable it to converge on accurate answers with fewer iterations.

\begin{figure}
    \centering
    \includegraphics[width=0.4\textwidth]{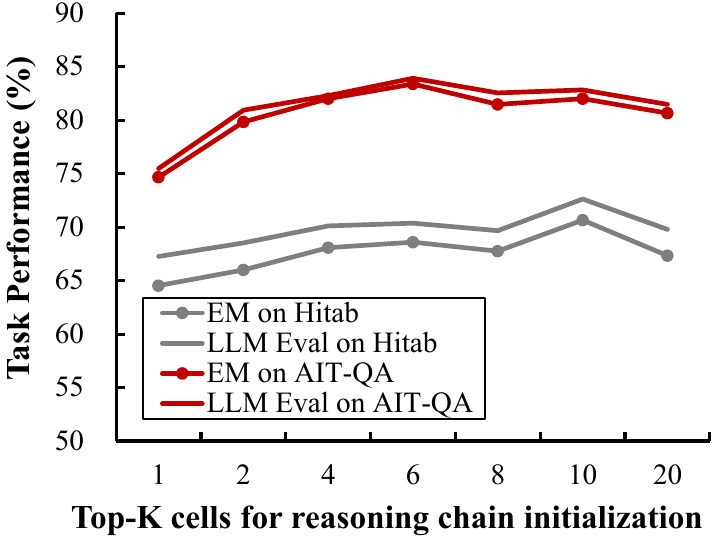}
    \caption{Impact of reasoning trace initialization (i.e., tuning top-k relevant cells to initialize the trace). Proper reasoning trace initialization ensures that \themodel begins its reasoning with enough information without being overwhelmed by irrelevant details.}
    \label{fig:hyparam}
    \vspace{-5mm}
\end{figure}

\textbf{\textit{What is the impact of reasoning trace initialization?} -- Proper reasoning trace initialization ensures that \themodel begins its reasoning with enough information to make accurate inferences without being overwhelmed by irrelevant details}. 
Considering \themodel instructs the LLM to select a small set of nodes relevant to the question $Q$ as the initial set for the reasoning trace, we analyze the impact of the number of these initial nodes on \themodel's performance, as shown in Figure \ref{fig:hyparam}. The results indicate that performance suffers when the number of nodes is either too small or too large. With too few nodes, \themodel may lack sufficient information, ultimately degrading overall performance. Conversely, excessive nodes may introduce irrelevant information, interfering with the LLM's reasoning process and negatively impacting performance. A potential solution to these issues could involve equipping \themodel with the ability to trace back and try new reasoning paths on the graph. However, given our focus on the initial development of graph reasoning for complex table QA, we leave this for future research. Empirically, we recommend using at most 4 to 8 initial nodes, and all experiments in this paper choose up to 8 nodes without further tuning.

\section{Conclusions}
Closely revolving around the complex table QA, we introduce \themodel, a novel method that explicitly establishes a reasoning process to pinpoint the correct answers.
\themodel converts the table into a graph and performs step-by-step reasoning with intermediate actions, leading to a clear and efficient path to the answer.
Our extensive evaluation demonstrates that \themodel outperforms existing implicit reasoning methods across various benchmark datasets and LLM backbones. This success can be attributed to \themodel's ability to filter out irrelevant information, focusing the reasoning process on the most pertinent data. The powerful synergy between graphs and explicit reasoning harnesses the flexibility of graph representations while mitigating their inherent complexity. We believe this may offer immense promise for accurately extracting answers from complex tables.

Beyond handling complex table structures, addressing complex user queries is crucial for the real-world effectiveness of TableQA systems. Therefore, future work will extend this method to multi-turn conversation scenarios, enabling active exploration and clarification of complex user questions \cite{zhang-etal-2024-clamber, chen-etal-2023-travel}.

\section*{Limitations}
\textbf{Iterative reasoning can be more expensive than one-shot methods}. Our method, like other iterative reasoning approaches, requires multiple interactions with the LLM to answer a single question, which can be more expensive than one-shot methods. While our method proves to be more efficient in terms of the number of iterations compared to other iterative methods (as shown in Figure \ref{fig:iterations}), our future work will focus on combining one-shot and iterative reasoning techniques to minimize unnecessary interactions with the LLM. This combination will allow us to achieve a balance between cost-effectiveness and task performance.

\noindent\textbf{More advanced reasoning actions on the graph}. This paper introduces four simple yet flexible reasoning actions, demonstrating their effectiveness in our experiments. However, to further enhance the performance of graph reasoning, future research will explore the introduction of more advanced actions. For instance, extending the actions related to accessing neighbors of specified nodes to include n-hop neighbors could provide the model with a longer-term perspective. However, as this study focuses on introducing graph reasoning to complex table QA for the first time, the exploration of advanced actions will be deferred to our future research.

\noindent\textbf{Potential sensitivity to prompts}. Similar to all other studies on prompting LLMs \cite{cheng2021hitab, zhao2023tableparser, E5}, our evaluation results could be influenced by the prompts. The impact of prompt design and prompt optimization represents a significant area for further research within the field of LLMs. Exploring the influence of different prompts and identifying optimal ones are crucial for maximizing the performance and effectiveness of LLM-based methods.

\section*{Acknowledgements}
This work was supported in part by the National Natural Science Foundation of China (No. 62272330); in part by the Fundamental Research Funds for the Central Universities (No. YJ202219); in part by the Science Fund for Creative Research Groups of Sichuan Province Natural Science Foundation (No. 2024NSFTD0035); in part by the National Major Scientific Instruments and Equipments Development Project of Natural Science Foundation of China under Grant  (No. 62427820)

\bibliography{custom}

\begin{thebibliography}{29}
\providecommand{\natexlab}[1]{#1}

\bibitem[{Adlakha et~al.(2023)Adlakha, BehnamGhader, Lu, Meade, and Reddy}]{adlakha2023LLMeval}
Vaibhav Adlakha, Parishad BehnamGhader, Xing~Han Lu, Nicholas Meade, and Siva Reddy. 2023.
\newblock Evaluating correctness and faithfulness of instruction-following models for question answering.
\newblock \emph{arXiv preprint arXiv:2307.16877}.

\bibitem[{Agrawal et~al.(2024)Agrawal, Vasania, and Tan}]{agrawal2024llmsperformstructuredgraph}
Palaash Agrawal, Shavak Vasania, and Cheston Tan. 2024.
\newblock \href {https://arxiv.org/abs/2402.01805} {Can llms perform structured graph reasoning?}
\newblock \emph{Preprint}, arXiv:2402.01805.

\bibitem[{Chen(2022)}]{fewshot}
Wenhu Chen. 2022.
\newblock Large language models are few (1)-shot table reasoners.
\newblock \emph{arXiv preprint arXiv:2210.06710}.

\bibitem[{Chen et~al.(2023)Chen, Jin, Huang, Liu, and Lei}]{chen-etal-2023-travel}
Yue Chen, Dingnan Jin, Chen Huang, Jia Liu, and Wenqiang Lei. 2023.
\newblock \href {https://doi.org/10.18653/v1/2023.emnlp-main.234} {{TRAVEL}: Tag-aware conversational {FAQ} retrieval via reinforcement learning}.
\newblock In \emph{Proceedings of the 2023 Conference on Empirical Methods in Natural Language Processing}, pages 3861--3872, Singapore. Association for Computational Linguistics.

\bibitem[{Chen et~al.(2021)Chen, Chen, Smiley, Shah, Borova, Langdon, Moussa, Beane, Huang, Routledge et~al.}]{chen2021finqa}
Zhiyu Chen, Wenhu Chen, Charese Smiley, Sameena Shah, Iana Borova, Dylan Langdon, Reema Moussa, Matt Beane, Ting-Hao Huang, Bryan Routledge, et~al. 2021.
\newblock Finqa: A dataset of numerical reasoning over financial data.
\newblock \emph{arXiv preprint arXiv:2109.00122}.

\bibitem[{Cheng et~al.(2022)Cheng, Dong, Wang, Jia, Guo, Gao, Han, Lou, and Zhang}]{cheng2021hitab}
Zhoujun Cheng, Haoyu Dong, Zhiruo Wang, Ran Jia, Jiaqi Guo, Yan Gao, Shi Han, Jian-Guang Lou, and Dongmei Zhang. 2022.
\newblock \href {https://doi.org/10.18653/v1/2022.acl-long.78} {{H}i{T}ab: A hierarchical table dataset for question answering and natural language generation}.
\newblock In \emph{Proceedings of the 60th Annual Meeting of the Association for Computational Linguistics (Volume 1: Long Papers)}, pages 1094--1110, Dublin, Ireland. Association for Computational Linguistics.

\bibitem[{Deng et~al.(2022)Deng, Lei, Zhang, Lam, and Chua}]{deng2022pacific}
Yang Deng, Wenqiang Lei, Wenxuan Zhang, Wai Lam, and Tat-Seng Chua. 2022.
\newblock Pacific: Towards proactive conversational question answering over tabular and textual data in finance.
\newblock In \emph{Proceedings of the 2022 Conference on Empirical Methods in Natural Language Processing}, pages 6970--6984.

\bibitem[{Jia et~al.(2023)Jia, Guo, Jin, Yan, Du, Ma, Stankovic, Lozajic, Zoranovic, Ilic et~al.}]{jia2023getpt}
Ran Jia, Haoming Guo, Xiaoyuan Jin, Chao Yan, Lun Du, Xiaojun Ma, Tamara Stankovic, Marko Lozajic, Goran Zoranovic, Igor Ilic, et~al. 2023.
\newblock Getpt: Graph-enhanced general table pre-training with alternate attention network.
\newblock In \emph{Proceedings of the 29th ACM SIGKDD Conference on Knowledge Discovery and Data Mining}, pages 941--950.

\bibitem[{Jiang et~al.(2024)Jiang, Zhou, Zhao, Song, Zhu, Zhu, and Wen}]{jiang2024kgagent}
Jinhao Jiang, Kun Zhou, Wayne~Xin Zhao, Yang Song, Chen Zhu, Hengshu Zhu, and Ji-Rong Wen. 2024.
\newblock Kg-agent: An efficient autonomous agent framework for complex reasoning over knowledge graph.
\newblock \emph{arXiv preprint arXiv:2402.11163}.

\bibitem[{Katsis et~al.(2021)Katsis, Chemmengath, Kumar, Bharadwaj, Canim, Glass, Gliozzo, Pan, Sen, Sankaranarayanan et~al.}]{katsis2021aitqa}
Yannis Katsis, Saneem Chemmengath, Vishwajeet Kumar, Samarth Bharadwaj, Mustafa Canim, Michael Glass, Alfio Gliozzo, Feifei Pan, Jaydeep Sen, Karthik Sankaranarayanan, et~al. 2021.
\newblock Ait-qa: Question answering dataset over complex tables in the airline industry.
\newblock \emph{arXiv preprint arXiv:2106.12944}.

\bibitem[{Liu et~al.(2024)Liu, Wang, and Chen}]{liu2023mixsc}
Tianyang Liu, Fei Wang, and Muhao Chen. 2024.
\newblock \href {https://doi.org/10.18653/v1/2024.naacl-long.26} {Rethinking tabular data understanding with large language models}.
\newblock In \emph{Proceedings of the 2024 Conference of the North American Chapter of the Association for Computational Linguistics: Human Language Technologies (Volume 1: Long Papers)}, pages 450--482, Mexico City, Mexico. Association for Computational Linguistics.

\bibitem[{Madras et~al.(2018)Madras, Pitassi, and Zemel}]{madras2018predict}
David Madras, Toni Pitassi, and Richard Zemel. 2018.
\newblock Predict responsibly: improving fairness and accuracy by learning to defer.
\newblock \emph{Advances in neural information processing systems}, 31.

\bibitem[{Ren et~al.(2024)Ren, Tang, Yin, Chawla, and Huang}]{ren2024llmforgraphsurvey}
Xubin Ren, Jiabin Tang, Dawei Yin, Nitesh Chawla, and Chao Huang. 2024.
\newblock A survey of large language models for graphs.
\newblock \emph{arXiv preprint arXiv:2405.08011}.

\bibitem[{Shang and Huang(2024)}]{shang2024llmforgraphsurvey2}
Wenbo Shang and Xin Huang. 2024.
\newblock A survey of large language models on generative graph analytics: Query, learning, and applications.
\newblock \emph{arXiv preprint arXiv:2404.14809}.

\bibitem[{Sun et~al.(2023)Sun, Xu, Tang, Wang, Lin, Gong, Shum, and Guo}]{sun2023thinkongraph}
Jiashuo Sun, Chengjin Xu, Lumingyuan Tang, Saizhuo Wang, Chen Lin, Yeyun Gong, Heung-Yeung Shum, and Jian Guo. 2023.
\newblock Think-on-graph: Deep and responsible reasoning of large language model with knowledge graph.
\newblock \emph{arXiv preprint arXiv:2307.07697}.

\bibitem[{Tailor et~al.(2024)Tailor, Patra, Verma, Manggala, and Nalisnick}]{tailor2024learning}
Dharmesh Tailor, Aditya Patra, Rajeev Verma, Putra Manggala, and Eric Nalisnick. 2024.
\newblock Learning to defer to a population: A meta-learning approach.
\newblock In \emph{International Conference on Artificial Intelligence and Statistics}, pages 3475--3483. PMLR.

\bibitem[{Wang et~al.(2023)Wang, Sun, Li, and Gao}]{wang2023boosting}
Jianing Wang, Qiushi Sun, Xiang Li, and Ming Gao. 2023.
\newblock Boosting language models reasoning with chain-of-knowledge prompting.
\newblock \emph{arXiv preprint arXiv:2306.06427}.

\bibitem[{Wang et~al.(2021)Wang, Mahajan, Danilevsky, and Rosenthal}]{wang2021scientificqa}
Nancy~XR Wang, Diwakar Mahajan, Marina Danilevsky, and Sara Rosenthal. 2021.
\newblock Semeval-2021 task 9: Fact verification and evidence finding for tabular data in scientific documents (sem-tab-facts).
\newblock \emph{arXiv preprint arXiv:2105.13995}.

\bibitem[{Wang et~al.(2024)Wang, Zhang, Li, Eisenschlos, Perot, Wang, Miculicich, Fujii, Shang, Lee et~al.}]{wang2024chainofable}
Zilong Wang, Hao Zhang, Chun-Liang Li, Julian~Martin Eisenschlos, Vincent Perot, Zifeng Wang, Lesly Miculicich, Yasuhisa Fujii, Jingbo Shang, Chen-Yu Lee, et~al. 2024.
\newblock Chain-of-table: Evolving tables in the reasoning chain for table understanding.
\newblock \emph{arXiv preprint arXiv:2401.04398}.

\bibitem[{Wu et~al.(2023)Wu, Hu, Qi, Bi, Ren, Xie, and Song}]{wu2023retrirewriteans}
Yike Wu, Nan Hu, Guilin Qi, Sheng Bi, Jie Ren, Anhuan Xie, and Wei Song. 2023.
\newblock Retrieve-rewrite-answer: A kg-to-text enhanced llms framework for knowledge graph question answering.
\newblock \emph{arXiv preprint arXiv:2309.11206}.

\bibitem[{Yao et~al.(2023)Yao, Zhao, Yu, Du, Shafran, Narasimhan, and Cao}]{react}
Shunyu Yao, Jeffrey Zhao, Dian Yu, Nan Du, Izhak Shafran, Karthik Narasimhan, and Yuan Cao. 2023.
\newblock React: synergizing reasoning and acting in language models (2022).
\newblock \emph{arXiv preprint arXiv:2210.03629}.

\bibitem[{Ye et~al.(2023)Ye, Hui, Yang, Li, Huang, and Li}]{ye2023dater}
Yunhu Ye, Binyuan Hui, Min Yang, Binhua Li, Fei Huang, and Yongbin Li. 2023.
\newblock Large language models are versatile decomposers: Decomposing evidence and questions for table-based reasoning.
\newblock In \emph{Proceedings of the 46th International ACM SIGIR Conference on Research and Development in Information Retrieval}, pages 174--184.

\bibitem[{Zhang et~al.(2024{\natexlab{a}})Zhang, Qin, Deng, Huang, Lei, Liu, Jin, Liang, and Chua}]{zhang-etal-2024-clamber}
Tong Zhang, Peixin Qin, Yang Deng, Chen Huang, Wenqiang Lei, Junhong Liu, Dingnan Jin, Hongru Liang, and Tat-Seng Chua. 2024{\natexlab{a}}.
\newblock \href {https://doi.org/10.18653/v1/2024.acl-long.578} {{CLAMBER}: A benchmark of identifying and clarifying ambiguous information needs in large language models}.
\newblock In \emph{Proceedings of the 62nd Annual Meeting of the Association for Computational Linguistics (Volume 1: Long Papers)}, pages 10746--10766, Bangkok, Thailand. Association for Computational Linguistics.

\bibitem[{Zhang et~al.(2024{\natexlab{b}})Zhang, Gao, and Lou}]{E5}
Zhehao Zhang, Yan Gao, and Jian-Guang Lou. 2024{\natexlab{b}}.
\newblock E5: Zero-shot hierarchical table analysis using augmented llms via explain, extract, execute, exhibit and extrapolate.
\newblock In \emph{Proceedings of the 2024 Conference of the North American Chapter of the Association for Computational Linguistics: Human Language Technologies (Volume 1: Long Papers)}, pages 1244--1258.

\bibitem[{Zhao et~al.(2023)Zhao, Ji, Zhang, He, Wang, Wang, Feng, and Zhang}]{zhao2023tableparser}
Bowen Zhao, Changkai Ji, Yuejie Zhang, Wen He, Yingwen Wang, Qing Wang, Rui Feng, and Xiaobo Zhang. 2023.
\newblock Large language models are complex table parsers.
\newblock \emph{arXiv preprint arXiv:2312.11521}.

\bibitem[{Zheng et~al.(2023)Zheng, Hao, Jiang, Lin, Lyu, She, and Wang}]{zheng2023tqa}
Mingyu Zheng, Yang Hao, Wenbin Jiang, Zheng Lin, Yajuan Lyu, Qiaoqiao She, and Weiping Wang. 2023.
\newblock Im-tqa: A chinese table question answering dataset with implicit and multi-type table structures.
\newblock In \emph{Proceedings of the 61st Annual Meeting of the Association for Computational Linguistics (Volume 1: Long Papers)}, pages 5074--5094.

\bibitem[{Zhong et~al.(2020)Zhong, ShafieiBavani, and Jimeno~Yepes}]{zhong2020imagetablereco}
Xu~Zhong, Elaheh ShafieiBavani, and Antonio Jimeno~Yepes. 2020.
\newblock Image-based table recognition: data, model, and evaluation.
\newblock In \emph{European conference on computer vision}, pages 564--580. Springer.

\bibitem[{Zhu et~al.(2021{\natexlab{a}})Zhu, Lei, Huang, Wang, Zhang, Lv, Feng, and Chua}]{zhu2021tatqa}
Fengbin Zhu, Wenqiang Lei, Youcheng Huang, Chao Wang, Shuo Zhang, Jiancheng Lv, Fuli Feng, and Tat-Seng Chua. 2021{\natexlab{a}}.
\newblock \href {https://doi.org/10.18653/v1/2021.acl-long.254} {{TAT}-{QA}: A question answering benchmark on a hybrid of tabular and textual content in finance}.
\newblock In \emph{Proceedings of the 59th Annual Meeting of the Association for Computational Linguistics and the 11th International Joint Conference on Natural Language Processing (Volume 1: Long Papers)}, pages 3277--3287, Online. Association for Computational Linguistics.

\bibitem[{Zhu et~al.(2021{\natexlab{b}})Zhu, Lei, Huang, Wang, Zhang, Lv, Feng, and Chua}]{zhu2021tat}
Fengbin Zhu, Wenqiang Lei, Youcheng Huang, Chao Wang, Shuo Zhang, Jiancheng Lv, Fuli Feng, and Tat-Seng Chua. 2021{\natexlab{b}}.
\newblock Tat-qa: A question answering benchmark on a hybrid of tabular and textual content in finance.
\newblock In \emph{Proceedings of the 59th Annual Meeting of the Association for Computational Linguistics and the 11th International Joint Conference on Natural Language Processing (Volume 1: Long Papers)}. Association for Computational Linguistics.

\end{thebibliography}


\appendix

\section{Implementation Details}
\label{thisis details}
Considering the financial issue, we experimented with two LLM backbones for all methods: Gemini-1.5 (\textit{gemini-1.5-flash-latest}) and Qwen2 (\textit{Qwen2-72B-Instruct}). To guarantee reproducibility, we set the same temperature and the seed for both backbones (if applicable). Importantly, to mirror real-world scenarios, \themodel is conducted without prior knowledge of the table headers.
For baseline methods, cells located at the top or left of the table were assumed to be headers. 

\subsection{Implementation of Baselines}

For all baselines, we used the official code from the original paper's GitHub repository or the official prompt from the original paper for implementation. We refer to Appendix \ref{prompt for baselines} for prompts.

Importantly, given a complex table in a document, we cannot definitively determine the location of headers beforehand. The flexible nature of complex tables allows headers to appear in various positions, not solely at the top or left edges. This necessitates alternative solutions beyond manual header cell annotation, as it is not a practical approach \cite{zheng2023tqa}. To this end, we aim to simulate this real-world scenario by simply removing the header cell information in the prompts of the TableParser \cite{zhao2023tableparser}.

\subsection{Implementation of LLM-based Evaluator}
\label{tas}
While exact match scores are commonly employed in evaluation \cite{zheng2023tqa}, the flexible nature of LLM outputs makes direct alignment or similarity measures, such as exact match, inadequate for assessing our method and baselines \cite{zhao2023tableparser}. Therefore, we employ an LLM as an evaluator, following the approach of \citet{zhao2023tableparser, adlakha2023LLMeval}, to verify the accuracy of predictions made by all methods. Specifically, we utilize Gemini 1.5 (\textit{gemini-1.5-flash-latest}) as the backbone of the LLM evaluator, considering cost-effectiveness. The prompt template used for this evaluation is presented in Appendix \ref{LLM eval}.

\subsection{Implementation of \themodel}

\subsubsection{Graph Representation Initialization}
\begin{itemize}[leftmargin=*]
    \item \textbf{Transforming Table to Graph}. \themodel transforms the input table into an undirected graph, where edges connect cells within the same row or column.

    \item \textbf{Reasoning Trace Initialization}. To filter out irrelevant information and initialize the reasoning trace, we guide the LLM to identify nodes relevant to the given user question, effectively constructing a subgraph. In our implementation, we employ gte-base (\url{https://huggingface.co/thenlper/gte-base}) to retrieve relevant nodes from the embedding space and the LLM itself to generate nodes through in-context learning. Finally, given the two sets of nodes obtained, we take the intersection and obtain at most 8 most relevant ones. See Figure \ref{fig:prompt-ours-selectcell} for prompts and the format of the reasoning trace is illustrated in Figure \ref{fig:prompt-graph-example}.
\end{itemize}

\subsubsection{Graph Reasoning}
We employ three processes to facilitate step-by-step reasoning over the graph via Thought-Action-Update. Prompts for the three-step process are provided in Figure \ref{fig:prompt-Action} and Figure \ref{fig:prompt-Thought}. During the reasoning, we defined four reasoning actions, acting as four functions. Here, we provide the implementation of these actions.
\begin{itemize}[leftmargin=*]
 \item \textit{VisitNode}(\textit{query}:Str). This function/action serves as an intermediary, taking query from the LLM and accessing the corresponding nodes within the graph. Due to the inherent flexibility of the LLM's output, its responses may not always directly align with the node names in the graph. To address this challenge, we implemented a hybrid approach that combines keyword matching and semantic retrieval, implemented by gte-base. This approach aims to identify the most accurate nodes based on the LLM's output, ensuring a robust connection between the LLM and the graph structure.

 \item \textit{GetAllNeighbours}(\textit{node}:Tuple). This function retrieves all neighbors of a specified node. Notably, we limit the LLM to specify the exact name of the visited node that it wants to explore. Moreover, our knowledge of the graph's structure allows us to programmatically and accurately determine the neighbors of any given node.

 \item \textit{GetSharedNeighbours}(\textit{node1}:Tuple, \textit{node2}:Tuple): This function returns the shared neighbors of two specified nodes in the graph. Same with the previous function, we can accurately identify and return the shared neighbors given our knowledge on the graph.

 \item \textit{AnswerQuestion}(): Answer the user question based on the available information. When the LLM selects this function, it signifies that the available information is sufficient to answer the question. At this point, we guide the LLM to generate the final answer using CoT prompts. The prompt template is depicted in Figure \ref{fig:prompt-Answer}.
\end{itemize}

\subsection{Implementation of Prompts}
\subsubsection{Prompts for Baselines}
\label{prompt for baselines}

\begin{itemize}
    \item \textit{One-shot implicit reasoning}. TableReasoner: Figure \ref{fig:prompt-TableReasoner}; TableParser: Figure \ref{fig:prompt-tableparser}.
    \item \textit{Iterative implicit reasoning}. We use their official codes from the Github.
    \item \textit{Chain-of-Table}. We use the official codes from the Github.
\end{itemize}

\subsubsection{Prompts for \themodel}
\begin{itemize}
    \item \textit{Reasoning trace initialization}. Figure \ref{fig:prompt-ours-selectcell}
    \item \textit{Graph Reasoning}. Figure \ref{fig:prompt-ours-system-inst}, Figure \ref{fig:prompt-Thought}, Figure \ref{fig:prompt-Action}, Figure \ref{fig:prompt-Answer}, and Figure \ref{fig:prompt-graph-example}.
\end{itemize}

\subsubsection{Prompts for LLM-based Evaluator}
\label{LLM eval}
We employ an LLM as an evaluator, following the approach of \citet{zhao2023tableparser, adlakha2023LLMeval}, to verify the accuracy of predictions made by all methods. The prompt template is illustrated in Figure \ref{fig:prompt-LLMeval}.

\section{Dataset Details and Statistics}\label{app:dataset statistics}
We consider two benchmark datasets of complex table QA, including AIT-QA and Hitab. 

Given a complex table in a document, we cannot definitively determine the location of headers beforehand. The flexible nature of complex tables allows headers to appear in various positions, not solely at the top or left edges. This necessitates alternative solutions beyond manual header cell annotation, as it is not a practical approach \cite{zheng2023tqa}. In this case, our experiments follows this real-world setting and removes the annotated table header information from the both datasets.

As for the AIT-QA, its header cells and data cells are stored separately in two-dimensional lists, recording only the content of the cells without preserving their positional information. An example of a table storage format is shown below. When the number of header cells in each column of the $column\_header$ and $row\_header$ lists is consistent, it is possible to reconstruct a table with an unknown header cell structure by following a top-to-bottom, left-to-right order. However, the number of header cells in the last column of the $column\_header$ list (5) is inconsistent with the number of header cells in other columns (2), making it impossible to determine the specific position of each header cell in the last column. Consequently, the reconstruction of a table with an unknown header cell structure is not feasible. Therefore, we removed tables like this where we cannot eliminate the influence of the known headers. Finally, the statistics for the test sets of the two datasets are presented in Table \ref{tab:dataset statisics}.

\begin{lstlisting}[language=json, basicstyle=\tiny]
# An example of a table storage format in AIT-QA
{
    "id": "tab-2", # The Table ID.
    "column_header":  # A list of column names in the table. Column names can be hierarchical and sublist captures the order of hierarchy.
    [
        [...],
        [
            "Increase (decrease) from 2017 (a):",
            "Latin"
	],
        [
            "Increase (decrease) from 2017 (a):",
            "Total"
        ],
        [
            "",
            "Increase (decrease) from 2017 (a):",
            "",
            "Domestic",
            "Atlantic"
        ]
    ], 
    "row_header":  # A list of row headers in the table. Row headers can be hierarchical and sublist captures the order of hierarchy.
    [
        [...],
        [
	    "Passenger load factor (points)"
        ],
        [
            "11"
        ]
    ],  
    "data":  # A list of rows. Each row is a list of row entries.
    [
        [...],
        [
            "",
            "",
            "",
            "",
            "",
            "(a) See Part II, Item 6, Selected Financial Data, of this report for the definition of these statistics."
	]
    ] 
}

\end{lstlisting}
\begin{table}[]
\renewcommand\arraystretch{1.2}
\centering
\resizebox{0.45\textwidth}{!}{%
    \begin{tabular}{lcccc}
        \toprule
        \multicolumn{1}{c}{\multirow{2}{*}{\textbf{Model}}} &
          \multicolumn{2}{c}{\textbf{Gemini 1.5}} &
          \multicolumn{2}{c}{\textbf{Qwen2}} \\ \cline{2-5} 
        \multicolumn{1}{c}{} &
          \multicolumn{1}{c}{\textbf{HRQ}} &
          \multicolumn{1}{c}{\textbf{HUQ}} &
          \multicolumn{1}{c}{\textbf{HRQ}} &
          \multicolumn{1}{c}{\textbf{HUQ}} \\ \hline
        \multicolumn{5}{l}{\textit{One-shot Implicit Reasoning}}                           \\
        TableReasoner  & \textbf{71.05} & 63.92          & 73.68          & 76.98          \\
        TableParser    & 63.16          & 63.23          & 65.79          & 64.60          \\ \hline
        \multicolumn{5}{l}{\textit{Iterative Implicit Reasoning}}                          \\
        ReAct          & 44.74          & 47.77          & 60.53          & 70.10          \\
        E5             & 61.84          & 58.76          & 52.63          & 57.39          \\
        MIX-SC         & 56.58          & \underline{ 73.45}    & \textbf{82.89} & \underline{ 84.54}    \\ \hline
        \multicolumn{5}{l}{\textit{Iterative Explicit Reasoning}}                          \\
        Chain-of-Table & 25.00          & 16.84          & 48.68          & 49.48          \\
        Ours           & \underline{ 69.74}    & \textbf{84.54} & \underline{ 81.58}    & \textbf{90.03} \\ \hline
\end{tabular}%
}
\caption{Detailed evaluation on different user questions, categorized by whether their answers are related to the table headers or not. "HRQ" represents "Header-related questions", and "HUQ" represents "Header-unrelated questions". Here, EM is used. }
\label{tab:aitqa-EM-result}
\end{table}

\begin{table}[t]
    \resizebox{0.5\textwidth}{!}{
        \small
        \setlength{\tabcolsep}{6pt}
        \renewcommand{\arraystretch}{1.5}
        \begin{tabular}{cccc}
            \toprule
            \textbf{Dataset} & \textbf{\makecell{Table \\number }} & \textbf{QA Pair}  & \textbf{Domain}\\
            \midrule
            \makecell{Hitab} & 538  & 1,584   & Open domain\\
            \makecell{AIT-QA} & 80 & 367  & Airline industry\\
            \bottomrule
        \end{tabular}%
    }
    \caption{Test set statistics for the two datasets.}
    \label{tab:dataset statisics}
\end{table}

\section{Case Study of \themodel Reasoning}
\label{case}
\begin{figure}
    \centering
    \includegraphics[width=0.5\textwidth]{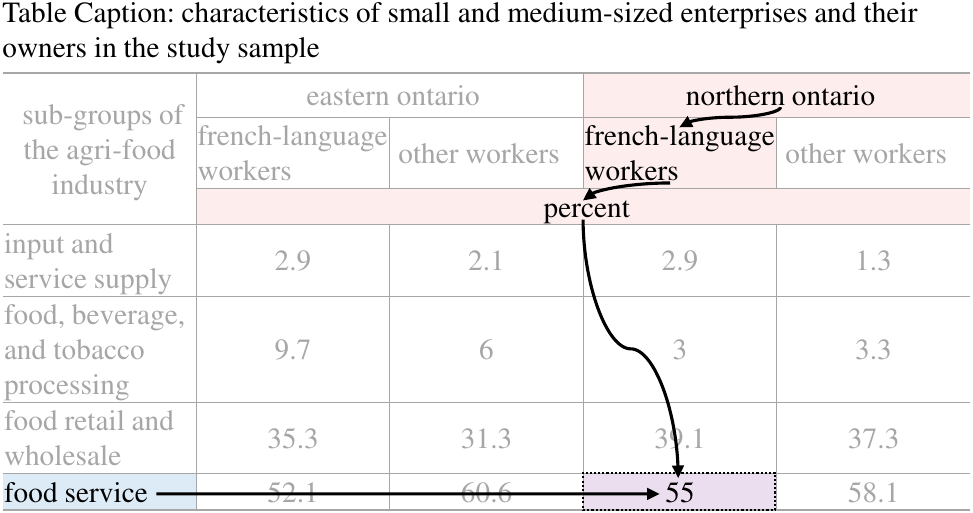}
    \caption{Illustration of the graph reasoning of \themodel. \textbf{User Question}: in northern ontario, what proportion of workers who have worked in the restaurant and food services sector was french-language workers?}
    \label{fig:case study}
\end{figure}

For better understanding, we involve a case study to reveal the reasoning process of \themodel. The examples of the user question and the complex table from the Hitab dataset are illustrated in Figure \ref{fig:case study}. The reasoning process of \themodel is provided in Table \ref{tab:ca}, which details each reasoning step. 

\textbf{User Question of Case Study}: in northern ontario, what proportion of workers who have worked in the restaurant and food services sector was french-language workers?


\section{Other Experimental Findings}
According to Table \ref{tab:main results}, \themodel demonstrates balanced performance across both Exact Match (EM) and LLM Evaluation metrics, unlike most baselines where LLM Evaluation scores significantly surpass EM, particularly for direct prompt methods. This balanced performance, achieved through \themodel's graph representation and reasoning framework, signifies the generation of answers that are both factually accurate and adhere to the desired format, a crucial factor in factoid question answering.

Additionally, Table \ref{tab:aitqa-EM-result} provides a detailed evaluation of the EM metric for different user questions, categorized based on whether their answers are related to the table headers. The results demonstrate that \themodel performs robustly across both question types, underscoring the effectiveness of our self-inferred headers.

\begin{table*}[]
    \centering
    \renewcommand\arraystretch{1.2}
    \resizebox{1.0\textwidth}{!}{%
    \begin{tabular}{p{2cm}p{10cm}p{10cm}}
        \toprule
        \textbf{Step ID} &
          \textbf{Reasoning Step or Action} &
          \textbf{Current Reasoning Trace} (in the form of subgraph) \\ \midrule
        0 &
          \begin{tabular}[c]{@{}p{10cm}@{}}\textit{Graph Representation Initialization} \\via finding cells that relevant to the user question.\end{tabular} &
          \begin{tabular}[c]{@{}p{10cm}@{}}{[}(1, 3, '\textcolor{blue}{french-language workers}'), (6, 0, '\textcolor{blue}{food service}'){]} has the following shared neighbors: {[}(1, 0, '\textcolor{blue}{sub-groups of the agri-food industry}'), (6, 3, '\textcolor{blue}{55.0}'){]}.\\ {[}(6, 0, '\textcolor{blue}{food service}'), (0, 3, '\textcolor{blue}{northern ontario}'){]} has the following shared neighbors: {[}(0, 0, '\textcolor{blue}{sub-groups of the agri-food industry}'), (6, 3, '\textcolor{blue}{55.0}'){]}.\\ {[}(0, 3, '\textcolor{blue}{northern ontario}'),SAME COLUMN,(1, 3, '\textcolor{blue}{french-language workers}'){]}\end{tabular} \\ \hline
        1 &
          \begin{tabular}[c]{@{}p{10cm}@{}}\textbf{Thought}: The question asks for the percentage of French-language workers in the "food services" sector in Northern Ontario. The graph includes both "Northern Ontario" and "Food Services," but the percentages are not provided, so we need to find them in the same column.\\
          \\\textbf{Action}: GetAllNeighbours((1, 3, "french-language workers"))\\\textit{Action Output}: \\ Same row: {[}(1, 0, 'sub-groups of the agri-food industry'), (1, 1, 'french-language workers'), (1, 2, 'other workers'), (1, 3, 'french-language workers'), (1, 4, 'other workers'){]}.\\ Same column: {[}(0, 3, 'northern ontario'), (1, 3, 'french-language workers'), (2, 3, 'percent'), (3, 3, '2.9'), (4, 3, '3.0'), (5, 3, '39.1'), (6, 3, '55.0'){]}. \\\\
          
          \textbf{Update}: No update
          
          \end{tabular} &
          Same as previous one \\ \hline
        2 &
          \begin{tabular}[c]{@{}p{10cm}@{}}\textbf{Thought}: We found the nodes for "Francophone Workers" and "Percentage" in the "Northern Ontario" column. Shared neighbors of 'Percentage' and 'Food Services' (6, 3, '55.0') suggest that 55.0 is the percentage of French-speaking workers in the 'Food Services' subgroup in Northern Ontario.\\ \\
          
          \textbf{Action}: VisitNode('Percentage')\\ \textit{Action Output}: None\\\\

          \textbf{Update}: Add node 'percent'

          \end{tabular} &
          \begin{tabular}[c]{@{}p{10cm}@{}}{[}(1, 3, 'french-language workers'), (6, 0, 'food service'){]} has the following shared neighbors: {[}(1, 0, 'sub-groups of the agri-food industry'), (6, 3, '55.0'){]}.\\ {[}(6, 0, 'food service'), (0, 3, 'northern ontario'){]} has the following shared neighbors: {[}(0, 0, 'sub-groups of the agri-food industry'), (6, 3, '55.0'){]}.\\ {[}(6, 0, '\textcolor{blue}{food service}'), (2, 3, '\textcolor{blue}{percent}'){]} has the following shared neighbors: {[}(2, 0, '\textcolor{blue}{sub-groups of the agri-food industry}'), (6, 3, '\textcolor{blue}{55.0}'){]}.\\ {[}(0, 3, 'northern ontario'),SAME COLUMN,(1, 3, 'french-language workers'),SAME COLUMN,(2, 3, 'percent'){]}          \end{tabular} \\ \hline
3 &
          \begin{tabular}[c]{@{}p{10cm}@{}}\textbf{Thought}: We found the nodes for "Francophone Workers" and "Percentage" in the "Northern Ontario" column. Shared neighbors of 'Percentage' and 'Food Services' (6, 3, '55.0') suggest that 55.0 is the percentage of French-speaking workers in the 'Food Services' subgroup in Northern Ontario.\\ \\
          
          \textbf{Action}: GetSharedNeighbours((2, 3, 'percent'), (6, 0, 'food service'))\\ \textit{Action Output}: {[}(2, 0, 'sub-groups of the agri-food industry'), (6, 3, '55.0'){]}\\\\

          \textbf{Update}: No update
          \end{tabular} & Same as previous one \\ \hline
          
        4 &
          \begin{tabular}[c]{@{}p{10cm}@{}}\textbf{Thought}: We found the nodes for "Francophone Workers" and "Percentage" in the "Northern Ontario" column. Shared neighbors of 'Percentage' and 'Food Services' (6, 3, '55.0') indicate that 55.0 is the percentage of French-speaking workers in the 'Food Services' subgroup in Northern Ontario.\\ \\
          \textbf{Action}: AnswerQuestion()\end{tabular} &
          Same as previous one \\ \hline
        5 &
          \textbf{Answer Generation}: 55.0 &
           \\ \bottomrule
    \end{tabular}%
}
\caption{Reasoning process of \themodel. \textit{User Question}: in northern ontario, what proportion of workers who have worked in the restaurant and food services sector was french-language workers? \textit{Answer}: 55.}
\label{tab:ca}
\end{table*}

\begin{figure*}[ht]
\begin{tcolorbox}[left=1mm,right=1mm,top=0.mm, bottom=0mm,colback=white]
\begin{lstlisting}[style=demo]
Read the table below regarding "{table_caption}" to answer the following questions.
The output format should be "The answer is xxx, yyy, zzz". If the answer contains multiple segments of text, please separate them with commas(",").

{table}

read the table first, and then answer the given question. 

Question: {question}
\end{lstlisting}
\end{tcolorbox}
\vspace{-3.5mm}
\caption{Prompt for TableReasoner }
\label{fig:prompt-TableReasoner}
\end{figure*}

\begin{figure*}[ht]
\begin{tcolorbox}[left=1mm,right=1mm,top=0.mm, bottom=0mm,colback=white]
\begin{lstlisting}[style=demo]
Suppose you are an expert in statistical analysis.
You will be given a table described in a special format.
Your task is to answer the questions based on the content of the table.

The table is described as follows:
1. The title means the title of the table.
2. We represent cell tuples as (C1, C2, C3), where C1 denotes the row, C2 denotes the column, and C3 denotes the content. 

For examples:
The tuple (7, 0, 416) represents a cell at row 7, column 0, with a value of 416. Make sure you read and understand these instructions carefully.

Let's think step by step as follows and give full play to your expertise as a statistical analyst: 
1. Clearly understand the question and the information needed to answer the question to determine the necessary information to extract. 
2. Have a comprehensive understanding of the data in the table, including the meaning, data types, and formats of each column and row tuples (Note: There are usually summative tuples in the table, such as all, combine, total, sum, average, mean, etc. These tuples help you skip a lot of operations).  
3. Perform statistical, calculation, sorting, grouping, or other operations on the tuples you selected before to extract useful information based on the question's requirements.

You MUST answer each question in the format below line by line (Note: Keep your answer concise): 
1. Cell: The cell tuples most relevant to the answer. 
2. Operation: the operation you performed on the tuples you selected. 
3. Answer: your answer (A number, noun, phrase, or set of data). 
And if the answer is not contained within the context, say "I don't know".

Title: {TABLE_TITLE_HERE}

Cells: {TABLE_NON_HEADER_HERE}
 
Q: {QUSTION_HERE}
A:
\end{lstlisting}
\end{tcolorbox}
\vspace{-3.5mm}
\caption{Prompt for TableParser }
\label{fig:prompt-tableparser}
\end{figure*}

\begin{figure*}[ht]
\begin{tcolorbox}[left=1mm,right=1mm,top=0.mm, bottom=0mm,colback=white]
\begin{lstlisting}[style=demo]
Suppose you are an expert in statistical analysis.
You will be given a table described in a special format. 
Your task is to answer the Question based on the content of the table. 

Graph Definition: We consider each cell in the table as a node in the graph, represented by the tuple (Row Index, Column Index, Cell Content). For example, (1, 0, "test1") represents the node in the 1st row and 0th column, with content "test1". The tuples (1, 0, "test1"), (1, 4, "test2"), and (3, 0, "test3") represent three nodes, where "test2" and "test1" have a SAME ROW relationship, "test3" and "test1" have a SAME COLUMN relationship, and "test1" is the shared neighbor of "test2" and "test3".
\end{lstlisting}
\end{tcolorbox}
\vspace{-3.5mm}
\caption{System instruction for \themodel's graph reasoning}
\label{fig:prompt-ours-system-inst}
\end{figure*}

\begin{figure*}[ht]
\begin{tcolorbox}[left=1mm,right=1mm,top=0.mm, bottom=0mm,colback=white]
\begin{lstlisting}[style=demo]
====================================  System Instruction ================================== 
Suppose you are an expert in statistical analysis.
You will be given a Table described in a special format.
Your task is to identify the cells in the Table that is most relevant to the Question.

Each cell in the Table is represented by a tuple (Row Index, Column Index, Cell Content). 
For example, the tuple (7, 0, "416") represents a cell at row 7, column 0, with a value of "416".
Make sure you read and understand these instructions carefully.

=========================================  Prompt ========================================= 
Let's think step by step as follows and give full play to your expertise as a statistical analyst: 
1. **Understand the Question**: Clearly understand the Question and the information needed to answer the Question to determine the necessary information to extract. 
2. **Analyze the Data Structure**: Have a comprehensive understanding of the data in the Table, including the meaning, data types, and formats of each cell tuples.    
3. **Select Relevant Data**: Based on the Question, identify the most relevant cell tuples. **Note:** Pay special attention to the header cell tuples in the Table, as they are often more relevant to the Question's semantics and can help in identifying the related evidence cell tuples.

{examples}

{table}

**Question:** {question}

Output format instructions:
1. Outputs cell tuples in descending order of relevance. 
2. Using this JSON schema: Tuple = {"tuple": tuple, "explanation": str}.  Return a `list[Tuple]`. 
\end{lstlisting}
\end{tcolorbox}
\vspace{-3.5mm}
\caption{Prompt and system instruction for \themodel's reasoning trace initialization. Subsequently, we identify the intersection of the results generated by this prompt and the results returned by the gte-base retriever.}
\label{fig:prompt-ours-selectcell}
\end{figure*}

\begin{figure*}[ht]
\begin{tcolorbox}[left=1mm,right=1mm,top=0.mm, bottom=0mm,colback=white]
\begin{lstlisting}[style=demo]
You can interact with the graph through two steps: "Thought" and "Action", to complete the question answering task step by step:
1. In the "Thought" step, thoroughly examine the question and the existing data. Determine if the current data is sufficient to answer the question:
    a. If the existing information is sufficient, proceed the "Action" step and call the "AnswerQuestion" function to give the answer.
    b. If more information is needed, call the functions in the 'Action' step to obtain useful information.
2. In the "Action" step, you can call the following functions to get more node information from the graph :
    a. VisitNode(query): Retrieve the node from the graph that is semantically closest to the keyword 'query'(given as a str). Note: 'query' cannot be the known Cell Content, to avoid meaningless calls;
    b. GetAllNeighbours(node): Get all neighboring nodes in the same row and column of the specified node (given as a tuple) from the graph.
    c. GetSharedNeighbours(node1, node2): Get all shared neighbors between two specified nodes (also represented as tuples) in the graph. 
    d. AnswerQuestion(): Answer the question based on the available information. 

Let's think step by step as follows and give full play to your expertise as a statistical analyst:
1. **Understand the Question**: Clearly understand the Question, clarify the relationships between the existing data, and organize the information needed to answer the Question.
2. **Analyze the Data Structure**: Have a comprehensive understanding of the data in the graph, including the meaning, data types, and formats of each nodes. **Note:** Pay special attention to some **summative or aggregated nodes** (e.g., "all", "combine", "total", "sum", "average", "mean",  "percent", "percentage", "proportion", "%", "probability", "likelihood", etc.), as these nodes help you skip a lot of operations.
3. **Select Relevant Data**: Based on the Question, identify the most relevant nodes. (**Note:** Some cells may have identical Cell Content, so avoid greedy searches when necessary; Focus on the most relevant and directly related data to the Question at hand.)
4. **Avoid Redundant Calculations**: Before performing any calculations or operations, first check if the needed information is already available in the graph. If so, directly use this information.

Table: {Table}

Question: {Question}

{Reasoning Trace}

{Interaction History}

Please integrate all the current information to output your Thought step {step}. The output format is: "Thought step {step}: {your thought}"
\end{lstlisting}
\end{tcolorbox}
\vspace{-3.5mm}
\caption{Thought Step Prompt of \themodel. Note that the Update Step does not rely on prompt.}
\label{fig:prompt-Thought}
\end{figure*}

\begin{figure*}[ht]
\begin{tcolorbox}[left=1mm,right=1mm,top=0.mm, bottom=0mm,colback=white]
\begin{lstlisting}[style=demo]
You can interact with the graph through two steps: "Thought" and "Action", to complete the question answering task step by step:
1. In the "Thought" step, thoroughly examine the question and the existing data. Determine if the current data is sufficient to answer the question:
    a. If the existing information is sufficient, proceed the "Action" step and call the "AnswerQuestion" function to give the answer.
    b. If more information is needed, call the functions in the 'Action' step to obtain useful information.
2. In the "Action" step, you can call the following functions to get more node information from the graph :
    a. VisitNode(query): Retrieve the node from the graph that is semantically closest to the keyword 'query'(given as a str). Note: 'query' cannot be the known Cell Content, to avoid meaningless calls;
    b. GetAllNeighbours(node): Get all neighboring nodes in the same row and column of the specified node (given as a tuple) from the graph.
    c. GetSharedNeighbours(node1, node2): Get all shared neighbors between two specified nodes (also represented as tuples) in the graph. 
    d. AnswerQuestion(): Answer the question based on the available information. 

Table: {Table}

Question: {Question}

{Reasoning Trace}

{Interaction History}

Based on the results of your previous Thought step {step}, output your Action Step and Explanation.
Using this JSON schema: ActionStep = {"Function" : {"function_name": str, "parameters": list[tuple] | list[str] }, "Explanation": str}. Return a `list[ActionStep]`.
\end{lstlisting}
\end{tcolorbox}
\vspace{-3.5mm}
\caption{Action Step Prompt of \themodel. Note that the Update Step does not rely on prompt.}
\label{fig:prompt-Action}
\end{figure*}

\begin{figure*}[ht]
\begin{tcolorbox}[left=1mm,right=1mm,top=0.mm, bottom=0mm,colback=white]
\begin{lstlisting}[style=demo]
Let's think step by step as follows and give full play to your expertise as a statistical analyst:
1. **Understand the Question**: Clearly understand the Question, clarify the relationships between the existing data, and organize the information needed to answer the Question. 
2. **Analyze the Data Structure**: Have a comprehensive understanding of the data in the graph, including the meaning, data types, and formats of each nodes. **Note:** Pay special attention to some **summative or aggregated nodes** (e.g., "all", "combine", "total", "sum", "average", "mean",  "percent", "percentage", "proportion", "%", "probability", "likelihood", etc.), as these nodes help you skip a lot of operations.  
3. **Select Relevant Data**: Based on the Question, identify the most relevant nodes.
4. **Avoid Redundant Calculations**: Before performing any calculations or operations, first check if the needed information is already available in the graph. If so, directly use this information.
5. **Synthesize the Answer**: Use the selected data to construct a clear and concise answer. Ensure that the final answer directly addresses the question, using the most relevant and accurate data from the graph.

You MUST answer each question step by step as follows(Note: Keep your answer concise): 
1. Cell: The nodes most relevant to the answer. 
2. Operation: the operation you performed on the nodes you selected. 
3. Explanation: your explanation.
4. Answer: your final answer.
And if you need to extract relevant Cell Content from the graph as answer, do not add any units, symbols, or other explanatory text. Ensure that the extracted Content matches the original Cell Content in the graph exactly.

Table: {table}

Question: {Question}

{Reasoning Trace}

{Interaction History}

Please integrate all the current information to output your answer.
Using this JSON schema: Answer = {"cells" : list[str], "operation": str, "explanation": str, "answer": list[str] }. Return a `Answer`.
\end{lstlisting}
\end{tcolorbox}
\vspace{-3.5mm}
\caption{Answer Generation Prompt of \themodel}
\label{fig:prompt-Answer}
\end{figure*}

\begin{figure*}[ht]
\begin{tcolorbox}[left=1mm,right=1mm,top=0.mm, bottom=0mm,colback=white]
\begin{lstlisting}[style=demo]
==================================== Reasoning Trace ====================================== 
The reasoning trace below displays the neighbor relationships of the nodes/steps from your interaction history. This trace will dynamically update during your interaction process.
**Reasoning Trace:**
[(4, 4, '37.7'), (0, 2, 'as a percent of total female population in each cma/municipality')] has the following shared neighbors: [(0, 4, 'largest visible minority group as a percent of total female population in each cma/municipality'), (4, 2, '66.5')].
[(4, 0, 'brampton'), (0, 4, 'largest visible minority group as a percent of total female population in each cma/municipality')] has the following shared neighbors: [(0, 0, 'municipalities'), (4, 4, '37.7')].
[(4, 0, 'brampton'), (0, 2, 'as a percent of total female population in each cma/municipality')] has the following shared neighbors: [(0, 0, 'municipalities'), (4, 2, '66.5')].
[(4, 0, 'brampton'),SAME ROW,(4, 4, '37.7')]
[(0, 4, 'largest visible minority group as a percent of total female population in each cma/municipality'),SAME COLUMN,(4, 4, '37.7')]
[(0, 2, 'as a percent of total female population in each cma/municipality'),SAME ROW,(0, 4, 'largest visible minority group as a percent of total female population in each cma/municipality')]

==================================  Interaction History =================================== 
The steps below represent your interaction history on the graph. Please refer to them when selecting your next function. 
Step 1:["GetSharedNeighbours((4, 0, 'brampton'), (0, 2, 'as a percent of total female population in each cma/municipality'))"]

Thought step 1: The question asks for the percentage of the total female population in Brampton, where South Asian women and girls were the largest visible minority group. The nodes (4, 4, '37.7') and (4, 0, 'brampton') are relevant to the answer, as well as the shared neighbors (0, 4, 'largest visible minority group as a percent of total female population in each cma/municipality') and (0, 2, 'as a percent of total female population in each cma/municipality'). The percentage of the total female population in Brampton is the information needed to answer the question.
Action Step 1:
    1. Function: GetSharedNeighbours((4, 0, 'brampton'), (0, 2, 'as a percent of total female population in each cma/municipality')), Explanation: To find the percentage of the total female population in Brampton, we need to get the shared neighbors of the nodes (4, 0, 'brampton') and (0, 2, 'as a percent of total female population in each cma/municipality').
Observation Step 1:
    1. The result of GetSharedNeighbours((4, 0, 'brampton'), (0, 2, 'as a percent of total female population in each cma/municipality')) is: [(4, 0, 'brampton'), (0, 2, 'as a percent of total female population in each cma/municipality')] has the following shared neighbors: [(0, 0, 'municipalities'), (4, 2, '66.5')].
\end{lstlisting}
\end{tcolorbox}
\vspace{-3.5mm}
\caption{A example of Reasoning Trace and Interaction History of \themodel}
\label{fig:prompt-graph-example}
\end{figure*}

\begin{figure*}[ht]
\begin{tcolorbox}[left=1mm,right=1mm,top=0.mm, bottom=0mm,colback=white]
\begin{lstlisting}[style=demo]
Instruction: You are CompareGPT, a machine to verify the correctness of predictions. Answer with only yes/no.

You are given a question, the corresponding ground-truth answer and a prediction from a model. Compare the "Ground-truth Answer" and the "Prediction" to determine whether the prediction correctly answers the question. 
All information in the Ground-truth Answer must be present in the Prediction, including numbers and dates. 
You must answer "no" if there are any specific details in the Ground-truth Answer that are not mentioned in the Prediction. 
There should be no contradicting statements in the Prediction. The Prediction may contain extra information. If the Prediction states something as a possibility, treat it as a definitive answer.

Question: {question}
Ground-truth Answer: {ground_truth}
Prediction:  {response}

CompareGPT response:
\end{lstlisting}
\end{tcolorbox}
\vspace{-3.5mm}
\caption{LLM Evaluation Prompt}
\label{fig:prompt-LLMeval}
\end{figure*}

\end{document}